\documentclass[sigconf]{acmart}
\copyrightyear{2019} 
\acmYear{2019} 
\setcopyright{acmcopyright}
\acmConference[SIGIR '19]{Proceedings of the 42nd International ACM SIGIR Conference on Research and Development in Information Retrieval}{July 21--25, 2019}{Paris, France}
\acmBooktitle{Proceedings of the 42nd International ACM SIGIR Conference on Research and Development in Information Retrieval (SIGIR '19), July 21--25, 2019, Paris, France}
\acmPrice{15.00}
\acmDOI{10.1145/3331184.3331208}
\acmISBN{978-1-4503-6172-9/19/07}

\usepackage{booktabs} 
\usepackage{CJK}
\usepackage{subfigure}
\usepackage{sistyle}
\SIthousandsep{,}
\usepackage{makecell}
\usepackage{multirow}
\usepackage{enumitem}
\usepackage{filecontents}
\usepackage{bbm}
\usepackage{algorithm}
\usepackage{algorithmic}
\usepackage{pgfplots}
\usetikzlibrary{patterns}

\makeatletter
\g@addto@macro\normalsize{%
  \abovedisplayskip 2.2pt plus1pt 
  \belowdisplayskip 3pt plus1pt
  \abovedisplayshortskip  3pt plus1pt%
  \belowdisplayshortskip  3pt plus1pt
}
\makeatother

\definecolor{c1}{RGB}{255,255,157}
\definecolor{c2}{RGB}{190, 235, 159}
\definecolor{c3}{RGB}{94,180,210}

\begin{document}

\title{Outline Generation: Understanding the Inherent Content Structure of Documents}
\author{Ruqing Zhang, Jiafeng Guo, Yixing Fan, Yanyan Lan, and Xueqi Cheng}
\affiliation{
  \institution{
    CAS Key Lab of Network Data Science and Technology, Institute of Computing Technology, \\ Chinese Academy of Sciences, Beijing, China\\
    University of Chinese Academy of Sciences, Beijing, China\\} 
}
\email{{zhangruqing,guojiafeng,fanyixing,lanyanyan,cxq}@ict.ac.cn}

\begin{abstract}
In this paper, we introduce and tackle the Outline Generation (OG) task, which aims to unveil the inherent content structure of a multi-paragraph document by identifying its potential sections and generating the corresponding section headings. 
Without loss of generality, the OG task can be viewed as a novel structured summarization task. 
To generate a sound outline, an ideal OG model should be able to capture three levels of coherence, namely the coherence between context paragraphs, that between a section and its heading, and that between context headings. The first one is the foundation for section identification, while the latter two are critical for consistent heading generation. In this work, we formulate the OG task as a hierarchical structured prediction problem, i.e., to first predict a sequence of section boundaries and then a sequence of section headings accordingly. We propose a novel hierarchical structured neural generation model, named HiStGen, for the task. Our model attempts to capture the three-level coherence via the following ways. First, we introduce a Markov paragraph dependency mechanism between context paragraphs for section identification. Second, we employ a section-aware attention mechanism to ensure the semantic coherence between a section and its heading. Finally, we leverage a Markov heading dependency mechanism and a review mechanism between context headings to improve the consistency and eliminate duplication between section headings. 
Besides, we build a novel W{\scriptsize IKI}OG dataset, a 
public collection which consists of over $1.75$ million document-outline pairs for research on the OG task.
Experimental results on our benchmark dataset demonstrate that our model can significantly outperform several state-of-the-art sequential generation models for the OG task.

\end{abstract}

%
%

\ccsdesc[500]{Computing methodologies~Natural language generation}

\keywords{Outline generation, Coherence, Hierarchical structured prediction}

\maketitle

\section{Introduction}

Document understanding is one of the critical and challenging tasks in information processing. There have been many related research topics in this direction, such as keyword detection \cite{liu2009unsupervised,tixier2016graph}, topic modeling \cite{hofmann1999probabilistic,blei2003latent}, headline generation \cite{dorr2003hedge,rush2015neural} and text summarization \cite{nallapati2016abstractive,see2017get}. Keyword detection and topic modeling aim to describe a document by a few important words or topics (i.e., distributions of words) for concise representation; While headline generation and text summarization attempt to compress the document into one or a few sentences to capture the key information. As we can see, most existing research on document understanding has focused on the coarse-grained understanding of documents by capturing its global semantics.
In this paper, we attempt to provide fine-grained understanding of documents by unveiling its inhere content structure \cite{langer2004text,stede2010identifying}, i.e., to understand how the document is organized and what it talks about in each part .

\begin{figure*}[t]
	\centering
		\includegraphics[scale=0.33]{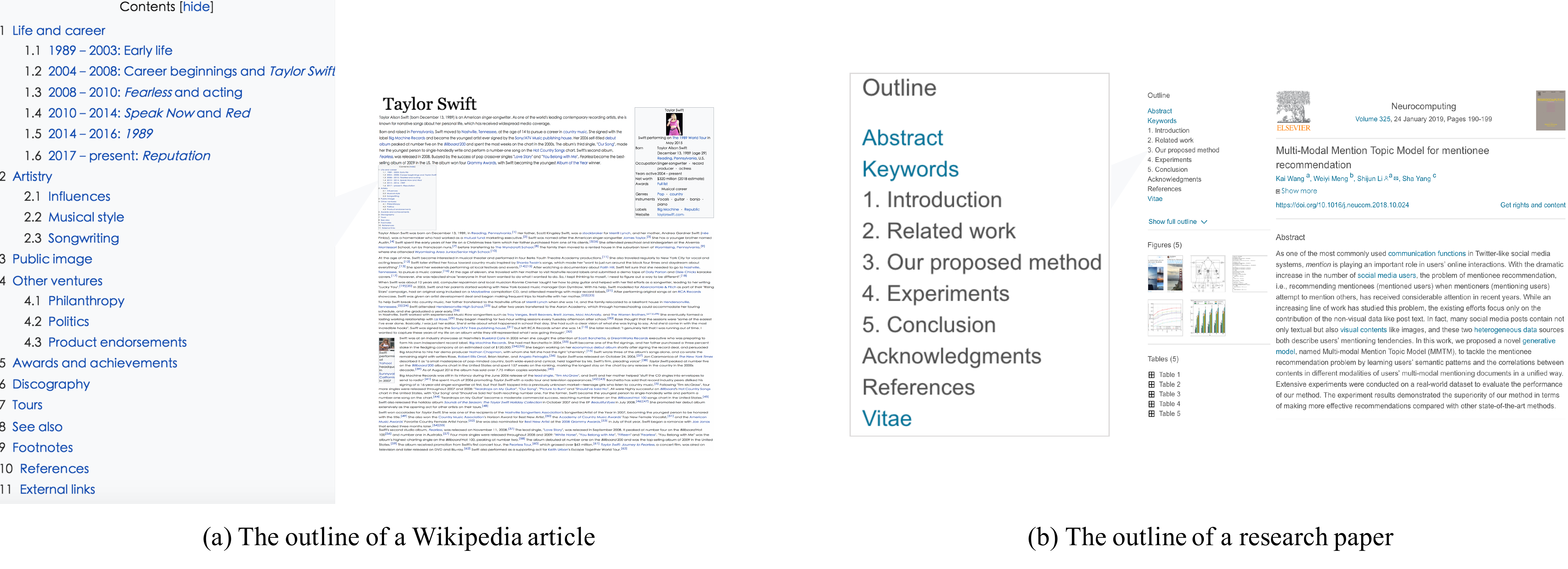}
		\caption{Examples of outlines in different types of documents.} 
\label{fig:outline}
\end{figure*}

We thus introduce the Outline Generation (OG) task in this work. Given a multi-paragraph document, the OG task aims to identify its potential sections and generate the corresponding section headings. Figure \ref{fig:outline} shows some typical outline of articles, where Figure \ref{fig:outline}(a) depicts the outline of a Wikipedia article\footnote{\url{https://en.wikipedia.org/wiki/Taylor_Swift}} with a two-level hierarchy, and Figure \ref{fig:outline}(b) depicts a typical outline of a research paper\footnote{\url{https://www.sciencedirect.com/science/article/pii/S0925231218312128}}.
As we can see, the outline can clearly capture the content structure of a document with concise text descriptions (i.e., section headings), which can not only help navigate the reading but also significantly reduce the cognitive burden over the document. Moreover, outlines can also facilitate a variety of text analysis applications such as text clustering and topic survey. 

In a conceptual level, the OG task could be viewed as a kind of summarization task. However, from the examples shown in Figure \ref{fig:outline}, we can find clear differences between the OG task and traditional summarization tasks. Firstly, the OG task produces a structured output with short descriptions (i.e., keywords or key phrases), while the output of traditional summarization is usually a set of unstructured sentences. Secondly, the OG task needs to summarize the paragraphs (into sections) in a strict sequential order, while the sentences in traditional summarization usually do not map to the paragraphs linearly. Thirdly, the section headings in one outline usually follow a similar style (e.g., topical headings as in Figure \ref{fig:outline}(a) and functional headings as in Figure \ref{fig:outline}(b)), while there is no such requirements in traditional summarization. Therefore, the OG task is actually a novel structured summarization task with its own special challenges.

If we take a further look at the OG task, we can find there are actually two structured prediction problem within it, i.e., to identify a sequence of sections (i.e., paragraphs with coherent information/topics), and to generate a sequence of section headings (i.e., short descriptions that summarize the sections) accordingly. Both problems are non-trivial. For section identification, it is unknown how many sections there are in a document. For section heading generation, headings should be able to reflect the section content in a consistent style. To achieve these two goals, an ideal OG model should be able to capture three levels of coherence, namely the coherence between context paragraphs, that between a section and its heading, and that between context headings. The first one is the foundation for section identification, while the latter two are critical for consistent heading generation.

In this work, we formulate the OG task as a hierarchical structured prediction problem and introduce a novel hierarchical structured neural generation model, named HiStGen, to solve it.
In this model, we view the section boundary prediction problem as a first-level sequential labeling process, and the section heading generation as a second-level structured prediction which depends on the predicted boundary labels from the lower level.
For section identification, we employ a Markov paragraph dependency mechanism to model the coherence in adjacent paragraphs to help decide the section boundaries. 
For section heading generation, we leverage a section-aware attention mechanism \cite{bahdanau2014neural} to allow the decoder to focus on the most informative content within a section for heading generation.
Furthermore, we introduce a Markov heading dependency mechanism and a review mechanism \cite{chen2018keyphrase} between context headings. The Markov heading dependency mechanism is used for modeling the consistency between adjacent headings, while the review mechanism is employed to avoid the repetition in the generated headings.

To facilitate the study and evaluation of the OG task, we build a new benchmark dataset based on Wikipedia articles. As we can see, in most multi-paragraph Wikipedia articles, human editors would segment the article into several sections and provide the outline as an overview of the content structure. Therefore, we can directly leverage these articles to build the benchmark. Specifically, we collect Wikipedia articles with outlines under ``celebrity", ``cities''  and ``music'' category, and obtain hundreds of thousands of articles respectively. 
We remove the outlines from Wikipedia articles to form the raw text input. The task is to recover the sections and section headings simultaneously. We call this benchmark dataset as W{\scriptsize IKI}OG.

For evaluation, we compare with several state-of-the-art methods to verify the effectiveness of our model. Empirical results demonstrate that outline generation for capturing the inherent content structure is feasible and our proposed method can outperform all the baselines significantly. We also provide detailed analysis on the proposed model, and conduct case studies to provide better understanding on the learned content structure.

The main contributions of this paper include:
\begin{itemize}[leftmargin=*]
\item We introduce a new OG task for fine-grained document understanding.
\item We propose a novel hierarchical structured neural sequential generation model to capture the three levels of coherence to solve the OG task.
\item We build a public W{\scriptsize IKI}OG dataset for the research of the OG task and conduct rigorous experiments to demonstrate the effectiveness of our proposed models.
\end{itemize}

\section{Related Work}

To the best of our knowledge, outline generation over a multi-paragraph document is a new task in the natural language processing community. The most closely related tasks to the OG task are keyword extraction, headline generation, text summarization and storyline generation tasks, which have been studied extensively in the past decades. 

Keyword extraction aims to automatically extract some keywords from a document. Most of the existing keyword extraction methods have
addressed this problem through two steps. The first step is to acquire a list of keyword candidates (e.g., n-grams or chunks) with heuristic methods \cite{hulth2003improved,shang2018automated}. The second step is to rank candidates on their importance to the document, either with supervised machine learning methods \cite{turney2002learning,frank1999domain,medelyan2009human,gollapalli2014extracting} or unsupervised machine learning  methods \cite{mihalcea2004textrank,wang2007keyword,bougouin2013topicrank,liu2009unsupervised}. However, these approaches could neither
identify keywords that do not appear in the text, nor capture the real semantic meaning behind the text. Recently, natural language generation models are used to automatically generate keywords. \citeauthor{meng2017deep} \cite{meng2017deep} applied an encoder-decoder framework \cite{cho2014learning} with a copy mechanism
\cite{gu2016incorporating} to this task, achieving state-of-the-art performance. \citeauthor{chen2018keyphrase} \cite{chen2018keyphrase} modeled correlation among multiple keywords in an end-to-end fashion to eliminate duplicate keywords and improve result coherence.

Headline generation aims to describe a document by a compact and informative headline, with the constraint that only a short sequence of words is allowed to generate \cite{dorr2003hedge}. Early work has pointed out that a purely extractive approach is not appropriate to generate headlines from the document text \cite{banko2000headline}.
This is due to two major reasons: (1) The single sentence extracted from the document is often longer than the desired headline size; (2) Sometimes the most important information is distributed across several sentences in the document. Hence, many studies have focused on either extracting and reordering n-grams from the document \cite{banko2000headline}, or selecting one or two informative sentences from the document, and then reducing them to the target headline size \cite{dorr2003hedge}. Recently, the task is formulated as a Seq2Seq learning problem and neural encoder-decoder architectures have been widely adopted to solve it. \citeauthor{lopyrev2015generating} \cite{lopyrev2015generating} trained an encoder-decoder recurrent neural network with attention for generating news headlines using the news articles from the English Gigaword corpus. \citeauthor{tanneural} \cite{tanneural} proposed to generate the headline from multiple summaries using a hierarchical attention model for the New York Times corpus. 

\begin{table}[t]
\renewcommand{\arraystretch}{1.2}
 \caption{A summary of key notations in this work.}
  \centering
  \begin{tabular}{p{1cm} | p{6cm}} \hline \hline
    $\mathcal{D}$ & A corpus of multi-paragraph documents \\ \hline 
     $d$ & A multi-paragraph document in $\mathcal{D}$\\\hline 
    $p_m$ & The $m$-th paragraph in $d$ \\\hline 
     $s_n$ & The $n$-th section in $d$ \\\hline  
     $y_n$ & The heading for the section $s_n$ \\\hline 
     $l_m$ & The section boundary label for $p_m$ \\ \hline 
     $w_{m,v}$ & The $v$-th word in $p_m$ \\\hline 
    $w_{n,u}$ & The $u$-th word in $y_n$ \\\hline 
     $V_m$ & The total number of words in $p_m$ \\\hline 
     $U_n$ & The total number of words in $y_n$ \\\hline
      $K_n$ & The total number of paragraphs in $s_n$ \\\hline  
     $M$ & The total number of paragraphs in $d$\\  \hline        
     $N$ & The total number of sections in $d$\\\hline   
    $g(\cdot)$ & The structured prediction model learnt for the OG task \\
    \hline\hline
  \end{tabular}
  \label{table:notations}
\end{table}

Text summarization is the process of automatically generating one or more natural summaries from an input document that retain the most important information. Most summarization models studied in the past are extractive in nature \cite{edmundson1964problems,radev1998generating,woodsend2010generation}, which try to extract the most important sentences in the document and rearranging them into a new summary. Recent abstractive summarization models have shown better flexibility and can generate more novel summaries. Many abstractive models \cite{xu2010keyword,rush2015neural,chopra2016abstractive} are based on the neural encoder-decoder architecture. To facilitate the research, a set of summarization tasks have been proposed in the Document Understanding Conference (DUC). These tasks often provide multiple human-generated reference summaries of the document for evaluation. 

\begin{figure*}[t]
	\centering
		\includegraphics[scale=0.38]{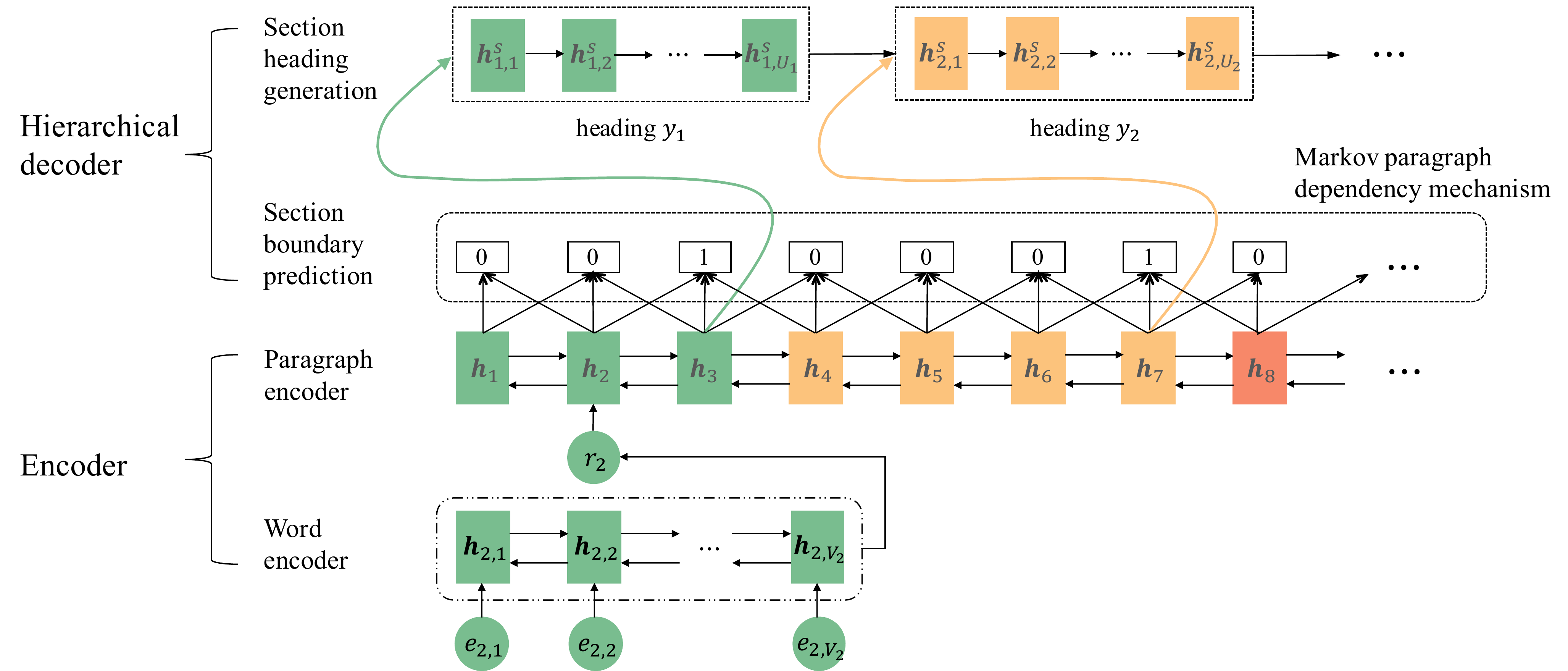}
		\caption{The basic architecture of hierarchical structured neural generation model (HiStGen). The detail of the section heading generation step in the hierarchical decoder is shown in Figure \ref{fig:chain}.}
\label{fig:HiStGen}
\end{figure*}

Storyline generation aims to summarize the development of certain events and understand how events evolve over time. \citeauthor{huang2013optimized} \cite{huang2013optimized} formalized different types of sub-events into local and global aspects. Some studies have been conducted in storyline generation with Bayesian networks to detect storylines \cite{zhou2015unsupervised,hua2016automatical}.
\citeauthor{lin2012generating} \cite{lin2012generating} firstly obtained relevant tweets and then generate storylines via graph optimization for the Tweets2011 corpus.

The OG task introduced in our work is related to the keyword extraction, headline generation, text summarization and storyline generation tasks but with some clear differences. Firstly, the output of keyword extraction is usually a set of unstructured keywords, while the OG task produces a structured output with short descriptions. Secondly, the output of the headline generation task is a single heading at the document-level with coarse-grained semantics, while the output of our OG task is a sequence of headings at the section-level with fine-grained semantics. Thirdly, text summarization aims to capture the major content of a document by producing a few unstructured sentences, while our OG task attempts to unveil the inherent content structure of a document by identifying its potential sections and generating the corresponding section headings. Finally, storyline generation is based on the multiple sub-events along the timeline, while the OG task focuses on the multiple sections. Therefore, most existing methods applied for these related tasks may not fit the OG task directly.

\begin{table}[t]
\renewcommand{\arraystretch}{1.2}
\caption{Data statistics: \#s denotes the number of sections, \#p denotes the number of paragraphs, and \#w denotes the number of words.}
  \centering
  \begin{tabular}{l | c| c| c} \hline \hline
      & celebrity & cities & music \\ \hline
      Wikipedia articles  & \num{568285} & \num{577698}  & \num{611162} \\
      Article vocabulary & \num{12265917} & \num{11514118} & \num{12629911} \\
      Outline vocabulary & \num{231107} & \num{204313} & \num{248935} \\
      Article avg \#s & 4.81 & 4.75 & 4.71 \\
      Article avg \#p & 39.70 & 37.34 & 39.83 \\ 
      Section avg \#p & 9.74 & 9.16 & 9.75  \\
      Heading avg \#w & 1.86 & 1.84 & 1.85 \\
    
      \hline
      \hline
  \end{tabular}
  
  \label{table:statistics}
\end{table}

\section{Problem Statement}

In this section, we introduce the OG task, and describe  the benchmark dataset W{\scriptsize IKI}OG in detail. A summary of key notations in this work is presented in Table \ref{table:notations}.

\subsection{Task Description}

Given a multi-paragraph document, the OG task aims to unveil its inherent content structure, i.e., to identify the potential sections (i.e., sequential paragraphs with coherent information/topics) of the document, as well as to generate the section headings (i.e., a short description that summarizes the section) correctly. Specifically, headings over different sections should be consistent in style and exclusive on topics, i.e., they should cover different aspects in a similar style. For example, as shown in Figure \ref{fig:outline} (b), headings in a research paper might include introduction, related work, method and so on. These headings are exclusive to each other and mainly describe the function of each section in the paper. 

Formally, given a document $d$ composed of a sequence of paragraphs $(p_1, p_2, \dots, p_M)$, the OG task is to learn a structured prediction model $g(\cdot)$ for $d$ to identify a sequence of sections $(s_1, s_2, \dots, s_N)$ and produce the corresponding section headings $(y_1,y_2,\dots,y_N)$ simultaneously,
\begin{equation}
g(p_1, p_2, \dots, p_M) =  (s_1, s_2, \dots, s_N; y_1,y_2,\dots,y_N), 
\end{equation}
where $M \ge N$.

\subsection{Benchmark Construction}

In order to study and evaluate the OG task, we build a new benchmark dataset W{\scriptsize IKI}OG. We take Wikipedia articles as our source articles since (1) Wikipedia is publicly available and easy to collect; (2) Most multi-paragraph Wikipedia articles contain outlines as an overview of the article, which are constructed by professional human editors. Specifically, we collect English Wikipedia\footnote{https://dumps.wikimedia.org/enwiki/} articles under three categories, i.e., ``celebrity'', ``cities'' and ``music''. We only make use of the first-level headings as our ground-truth, and leave the deeper-level headings (e.g., second-level headings) generation for the future study. Articles with no headings or more than ten first-level headings are removed, leaving us roughly $1.75$ million articles in total. Table \ref{table:statistics} shows the overall statistics of our W{\scriptsize IKI}OG benchmark dataset. 

For the OG task, we remove the outlines from Wikipedia articles, and concatenate all the paragraphs together to form the raw text input $\{p_m\}_{m=1}^{M}$. We record all the sections by their boundaries $\{l_m\}_{m=1}^{M}$ as well as all the corresponding section headings $\{y_n\}_{n=1}^{N}$. In this way, we obtain the $\scriptsize{<}$paragraph, section boundary label, section heading$\scriptsize{>}$ triples, i.e.,  $(\{p_m\}_{m=1}^{M}, \{l_m\}_{m=1}^{M}, \{y_n\}_{n=1}^{N})$, as ground-truth data for training/validation/testing.

\section{Our Approach}

In this section, we introduce our proposed approach for the OG task in detail. We first give an overview of the problem formulation and the model architecture. We then describe each component of our model as well as the learning procedure specifically.

\subsection{Overview}

Without loss of generality, the OG task can be decomposed into two structured prediction problems: 1) Section Identification: a sequential labeling process to identify the section boundaries; and 2) Section Heading Generation: a sequential generation process to produce short text descriptions for each identified section.
These two structured prediction problems are coupled in the sense that the section heading prediction is dependent on the section prediction results. Therefore, in this work, we formulate the OG task as a hierarchical structured prediction problem and introduce a novel hierarchical structured neural generation model, named HiStGen for short, to solve it. The overall architecture of HiStGen is illustrated in Figure \ref{fig:HiStGen}.

Basically, the HiStGen employs the encoder-decoder framework.
In the encoding phase, to obtain the representation of a multi-paragraph document, HiStGen utilizes the hierarchical encoder framework \cite{li2015hierarchical} to obtain the document representation.
The decoding phase is hierarchical, where we exploit three-level coherence for better OG prediction. Specifically, we employ a Markov paragraph dependency mechanism between context paragraphs for the section boundary prediction problem. Moreover, HiStGen employs a section-aware attention mechanism between a section and its heading, and a Markov heading dependency mechanism and a review mechanism between context headings for the heading generation problem whenever a new section is identified. We will discuss the details of these model designs in the following sections.

\subsection{Encoder}
\label{encoder}

The goal of the encoder is to map the input document to a vector representation.
In HiStGen, we adopt a hierarchical encoder framework, where we use a word encoder to encode the words of a paragraph $p_m$, and use a paragraph encoder to encode the paragraphs of a document $d$. 

As depicted in Figure \ref{fig:HiStGen}, each word $w_{m,v}$ in each paragraph $p_m$ is represented by its distributed representation $\textbf{e}_{m,v}$.
We use a bi-directional GRU as both the word and paragraph encoder, which summarizes not only the preceding words/paragraphs, but also the following words/paragraphs. 
The forward GRU in word encoder reads the words in the $m$-th paragraph $p_m$ in the left-to-right direction, resulting in a sequence of hidden states $(\overrightarrow{\textbf{h}}_{m,1}, \dots, \overrightarrow{\textbf{h}}_{m,V_m})$. 
The backward GRU reads $p_m$ in the reversed direction and outputs $(\overleftarrow{\textbf{h}}_{m,1}, \dots, \overleftarrow{\textbf{h}}_{m,V_m})$. 
We obtain the hidden state for a given word $w_{m,v}$ by concatenating the forward and backward hidden states, i.e., $\textbf{h}_{m,v} = [\overrightarrow{\textbf{h}}_{m,v}||\overleftarrow{\textbf{h}}_{m,v}]$. 
Then, we concatenate the last hidden states of the forward and backward passes as the embedding representation of the paragraph $p_m$, denoted as $\textbf{r}_m = [\overrightarrow{\textbf{h}}_{m,V_m}||\overleftarrow{\textbf{h}}_{m,1}]$. 
A paragraph encoder is used to sequentially receive the embeddings of paragraphs $\{\textbf{r}_m\}_{m=1}^{M}$ in a similar way. The hidden representation of each paragraph is given by $\textbf{h}_m = [\overrightarrow{\textbf{h}}_{m}||\overleftarrow{\textbf{h}}_{m}]$, where $\overrightarrow{\textbf{h}}_{m}$ and $\overleftarrow{\textbf{h}}_{m}$ are the forward and backward hidden states of the paragraph encoder respectively.

\subsection{Hierarchical Decoder}
The goal of the hierarchical decoder is to produce an outline for an input article, which could be decomposed into two dependent steps: (1) Section Boundary Prediction: to predict a sequence of section boundary labels over the paragraphs; and (2) Section Heading Generation: to generate the section heading for a newly detected section.

\subsubsection{Section Boundary Prediction}
This step is to break up a multi-paragraph document $\{p_1,p_2,\dots,p_M\}$ into multiple successive sections $\{s_1,s_2,\dots,s_N\}$ by predicting the section boundary labels $\{l_1,l_2,\dots,l_M\}$, where $l_m \in \{0,1\}$. If $l_m=0$, $p_m$ is the inner paragraph of a section and the section prediction continues.
If $l_m=1$, $p_m$ is the last paragraph of one section and the corresponding heading should be generated.  
Note that a section is a sequence of information coherent paragraphs, while the coherence modeling is non-trivial in nature. In this paper, we introduce a Markov paragraph dependency mechanism for modeling the coherence between context paragraphs and identifying section boundaries.

\begin{itemize}[leftmargin=*]
\item \textbf{Markov Paragraph Dependency Mechanism}. 
The key assumption of the Markov paragraph dependency mechanism is that the coherence between paragraphs has a Markov property. Therefore, we can identify a section, i.e., to decide whether a target paragraph is a last paragraph of a section, by looking at its previous and successive paragraph. 
As shown in Figure \ref{fig:HiStGen},  we utilize the hidden representation of the current paragraph $p_m$, the previous paragraph $p_{m-1}$, and the next paragraph $p_{m+1}$ to predict the section boundary label $l_m$. Specifically, the section boundary label $l_m$ is modeled with binary output:
\begin{equation}
\label{equ:boundary1}
p(l_m=1|p_{\le m}) = \sigma(\textbf{h}_{m-1} \cdot \textbf{W}_1 \cdot \textbf{h}_{m} + \textbf{h}_{m} \cdot \textbf{W}_2 \cdot \textbf{h}_{m+1}), 
\end{equation}
\begin{equation}
\label{equ:boundary2}
p(l_m=0|p_{\le m}) = 1 - \sigma(\textbf{h}_{m-1} \cdot \textbf{W}_1 \cdot \textbf{h}_{m} + \textbf{h}_{m} \cdot \textbf{W}_2 \cdot \textbf{h}_{m+1}), 
\end{equation}

where $\sigma(\cdot)$ stands for the sigmoid function, $\sigma(x) = \frac{1}{exp(-x)+1}$, and $\textbf{W}_1, \textbf{W}_2$ are learned parameters.
\end{itemize}

\begin{figure*}[t]
	\centering
		\includegraphics[scale=0.33]{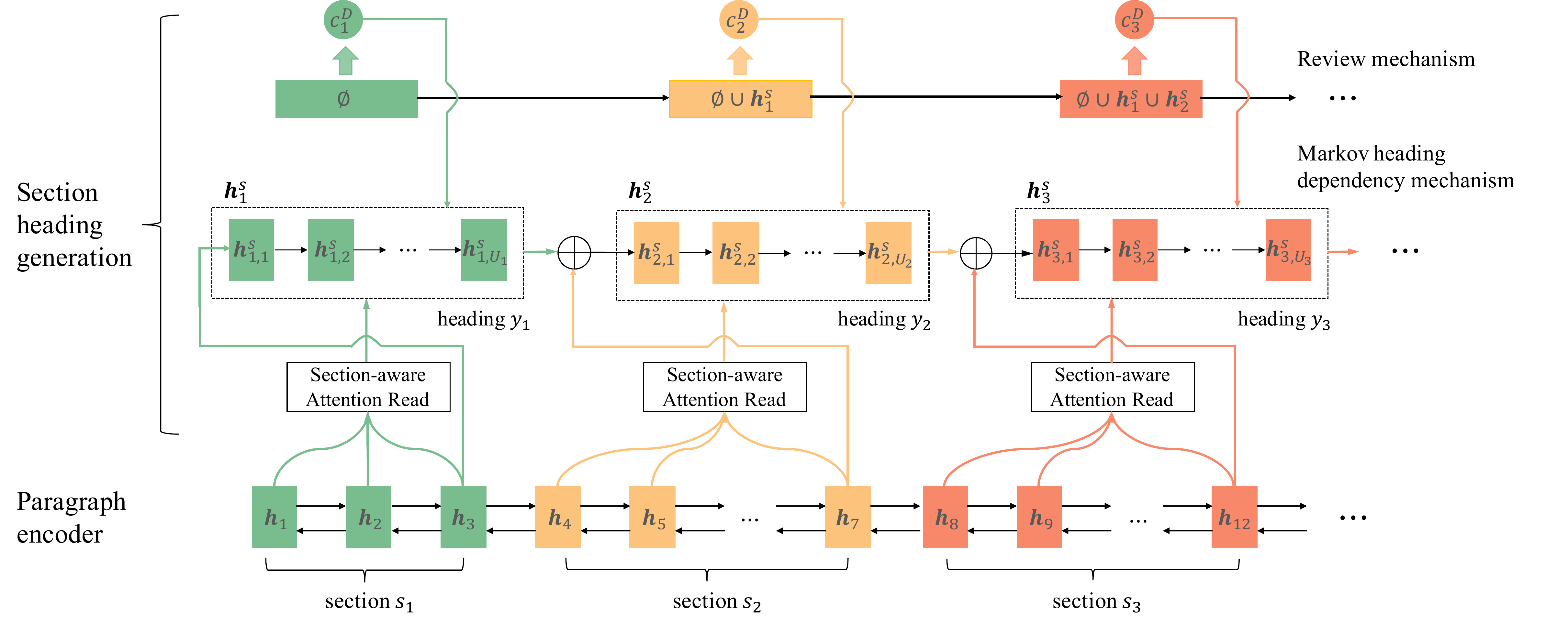}
		\caption{The detail of the section heading generation step in the HiStGen model.}
\label{fig:chain}
\end{figure*}

\subsubsection{Section Heading Generation}
\label{attention}
This step executes when a new section is detected, i.e., $l_m=1$. Based on the detected section $s_n$, to generate the heading $y_n$, we employ  
1) a section-aware attention mechanism: maintaining a section-aware context vector to make sure more important content in the target section is attended; 
2) a Markov heading dependency mechanism: maintaining the representation of the previously generated heading for new heading generation to improve the consistency between headings;
and 3) a review mechanism: maintaining a heading-aware context vector to utilize contextual information of generated headings to eliminate duplication between headings.  
The first one is used to capture the coherence between a section and its heading, and the latter two are used to capture the coherence between context headings.

Afterwards, the section-aware context vector $\textbf{c}_{n,u}$ and the heading-aware context vector $\textbf{c}_{n,u}^{D}$ are provided as extra inputs to derive the hidden state $\textbf{h}_{n,u}^{s}$ of the $u$-th word $w_{n,u}$ in $y_n$, and later the probability distribution for choosing the word $w_{n,u}$. 

Concretely, $\textbf{h}_{n,u}^{s}$ is defined as
\begin{equation}
\label{h_ts}
\textbf{h}_{n,u}^{s} = f_s(w_{n,u-1},\textbf{h}_{n,u-1}^{s},\textbf{c}_{n,u}^{D},\textbf{c}_{n,u}),
\end{equation}
where $f_s$ is a GRU unit, $w_{n,u-1}$ is the predicted word from vocabulary at $u-1$-th step when decoding the heading $y_n$. The probability distribution for choosing the word $w_{n,u}$ is defined as
\begin{equation}
\label{word_choose}
p(w_{n,u}|w_{\le n,<u},d) = f_g(w_{n,u-1}, \textbf{h}_{n,u}^{s},\textbf{c}_{n,u}^{D}, \textbf{c}_{n,u}),
\end{equation}
where $f_g$ is a nonlinear function that computes the probability vector for all legal output words at each output time. We now describe the specific mechanism in the follows.

\begin{itemize}[leftmargin=*]
\item \textbf{Section-Aware Attention Mechanism}. The key idea of the section-aware attention mechanism is to make the generation of a section heading focusing on the target section. 
Concretely, as shown in Figure \ref{fig:chain}, we maintain a section-aware context vector $\textbf{c}_{n,u}$ for generating the $u$-th word $w_{n,u}$ in the $n$-th heading $y_n$.
Based on the $n$-th section $s_n$,  $\textbf{c}_{n,u}$ is a weighted sum of the hidden representations of all the paragraphs in $s_n$: 
\begin{equation}
\label{equ:section-aware}
\textbf{c}_{n,u} = \sum_{k=1}^{K_n}\alpha_{u,k} \textbf{h}_{K_1+K_2+\dots,K_{n-1}+k},
\end{equation}
where $\alpha_{u,k}$ indicates how much the $k$-th paragraph $p_k$ from the source section $s_n$ contributes to generating the $u$-th word in target heading $y_n$, and is usually computed as:
\begin{equation}
\label{att-weight}
\alpha_{u,k} = \frac{\exp(\textbf{h}_{K_1+K_2+\dots,K_{n-1}+k} \cdot \textbf{h}_{n,u-1}^s)}{\sum_{i=1}^{K_n} \exp( \textbf{h}_{K_1+K_2+\dots,K_{n-1}+i} \cdot \textbf{h}_{n,u-1}^s)},
\end{equation}
where $\textbf{h}_{n,u-1}^s$ represents the hidden state (just before emitting the $u$-th word $w_{n,u}$ in $n$-th heading $y_n$) of the decoder.

\item \textbf{Markov Heading Dependency Mechanism}. The headings in an outline should be consistent in style and it is necessary to capture the dependence between context headings. To achieve this purpose, we introduce a Markov heading dependency mechanism, for the section heading generation process. Note that different from the Markov paragraph dependency mechanism, the Markov heading dependency mechanism only looks at the previous generated heading since there is no successive heading generated yet.

Concretely, as shown in Figure \ref{fig:chain}, 
the Markov heading dependency mechanism uses the accumulation of all the hidden states of the previous decoder and pass it to the next decoder. In this way, the generation of a new heading is decided by both the section content and the previous generated heading.

As we can see, the Markov heading dependency mechanism conveys strong dependency requirement between headings by involving all the previous states. The initial hidden state of the decoder $\textbf{h}_{n,0}^{s}$ of heading $y_n$ is the ``mixture'' of probabilities:
\begin{equation} 
\label{equ:aggregated} 
\textbf{h}_{n,0}^{s} = 
\left\{
             \begin{array}{l}  
             \textbf{W}_3 \cdot \textbf{z}_{n-1}  +  \textbf{W}_4 \cdot \textbf{h}_m + \textbf{b}_1,~~n > 1, \\  
              \textbf{W}_4  \cdot \textbf{h}_m + \textbf{b}_1,~~n=1 ,\\     
             \end{array}  
\right.
\end{equation}
where $\textbf{W}_3, \textbf{W}_4, \textbf{b}_1$ are learned parameters. 
$\textbf{h}_m$ is the representation of paragraph $p_m$, where $p_m$ is the last paragraph of the section $s_n$. The passed information $\textbf{z}_{n-1}$ is the average of all the output states of the decoder for the heading $y_{n-1}$ and defined as: 
\begin{equation}
\textbf{z}_{n-1} = \frac{1}{U_{n-1}} \sum_{u=1}^{U_{n-1}} \textbf{h}_{n-1, u}^s,
\end{equation}
where $\textbf{h}_{n-1, u}^s$ is the output state of the decoder for the heading $y_{n-1}$ at the $u$-th step.

\item \textbf{Review Mechanism}. Headings should cover all topics in the source document and be exclusive to each other. To avoid duplicate generation, we incorporate a review mechanism \cite{chen2018keyphrase} between context headings as shown in Figure \ref{fig:chain}. It models the correlation between the headings that have been generated and the heading that is going to be generated to generate a heading to cover topics that have not been summarized by previous headings.

Specifically, we construct a heading-aware review set which contains contextual information of generated headings. The heading-aware review set is defined as $R_{n,u}=
\{ \textbf{h}_{1,1}^{s},\textbf{h}_{1,2}^{s},\dots,\textbf{h}_{n,u-1}^{s}\}$, which is the collection of all the decoder hidden states before generating the $u$-th word $w_{n,u}$ in the $n$-th heading $y_n$. When decoding the word $w_{n,u}$, the heading-aware review set $R_{n,u}$ is integrated into the heading-aware context vector $\textbf{c}_{n,u}^{D}$:
\begin{equation}
\textbf{c}_{n,u}^{D} = \sum_{i,j=1}^{n,u-1}\beta_{nu}^{ij} \textbf{h}_{i,j}^{s},
\end{equation}
where $\beta_{nu}^{ij}$ indicated how much the $j$-word in the $i$-th heading contributed to generating the $u$-th word in target heading $y_n$, and is computed as:
\begin{equation}
\beta_{nu}^{ij} = \frac{\exp(\textbf{e}_{nu}^{ij})}{\sum_{p,q=1}^{n,u-1}\exp(\textbf{e}_{nu}^{pq})},
\end{equation}
where $\textbf{e}_{nu}^{ij}$ is defined as
\begin{equation}
\textbf{e}_{nu}^{ij} = v^{T}\tanh(\textbf{W}_5 \textbf{h}_{i,j}^{s} + \textbf{W}_6 \textbf{h}_{n,u-1}^{s} + \textbf{b}_2),
\end{equation}
where $\textbf{W}_5,\textbf{W}_6,\textbf{b}_2$ are learned parameters. The heading-aware review set gets updated consequently as $R_{n,u+1}=R_{n,u}\cup\{\textbf{h}_{n,u}^s\}$ in the decoding process.

\end{itemize}

\subsection{Model Training and Testing}

In the training phase, we employ maximum likelihood estimation (MLE) to learn our HiStGen model in an end-to-end way. Specifically, the training objective is a probability over the training corpus $\mathcal{D}$ with decomposition into the ordered conditionals:
\begin{equation}
\begin{aligned}
\arg \max_{\theta,\beta} & \sum_{d\in \mathcal{D}} \sum_{n=1}^{N}\log p(y_n|s_n,y_{<n};\theta) p(s_n|d;\beta).
\end{aligned}
\end{equation}

We apply stochastic gradient decent method Adam \cite{kingma2014adam} to learn the model parameters $\theta$ and $\beta$. 
Note that, during the training, we provide the model with the specific 
section boundary label $l_m$, and thus we do not have to sample.

In the testing phase, given a new multi-paragraph document, we compute Eqn. (\ref{equ:boundary1}) and (\ref{equ:boundary2}) to predict the section boundary label for each paragraph, and then pick the word with the highest probability using Eqn. (\ref{word_choose}) to generate the heading for each identified section.

\section{Experiments}

In this section, we conduct experiments to verify the effectiveness of our proposed model.

\subsection{Experimental Settings}

To evaluate the performance of our model, we conducted experiments on our W{\scriptsize IKI}OG benchmark dataset.
In preprocessing, all the words in documents and headings are white-space tokenized and lower-cased, and pure digit words and non-English characters are removed. Beyond the three separate datasets (i.e., ``celebrity'', ``cities'' and ``music''), we also mix them together to form a ``mixture'' dataset. For each dataset in W{\scriptsize IKI}OG, we randomly divide it into a training set (80\%), a development set (10\%), and a test set (10\%).

We construct two separate vocabularies for input documents and target headings by using $\num{130000}$ and $\num{16000}$ most frequent words on each side in the training data. All the other words outside the vocabularies are replaced by a special token $\scriptsize{<}$UNK$\scriptsize{>}$ symbol. We implement our models in Tensorflow\footnote{https:/www.tensorflow.org/}. Specifically, we use a bi-directional GRU for the word/paragraph encoder respectively and another forward GRU for the heading decoder, with the GRU hidden unit size set as 300 in both the encoder and decoder. The dimension of word embeddings in documents and headings is 300. The learning rate of Adam algorithm is set as $0.0005$. The learnable parameters (e.g., the parameters $W_1$, $W_2$ and $b_1$) are uniformly initialized in the range of $[-0.08, 0.08]$. The mini-batch size for the update is set as 64. We clip the gradient when its norm exceeds 5.

We run our model on a Tesla K80 GPU card, and we run the training for up to \num{12} epochs, which takes approximately two days. We select the model that achieves the lowest perplexity on the development set, and report results on the test set.

\subsection{Baselines}

\subsubsection{Model Variants}

Here, we first employ some degraded HiStGen models to investigate the effect of our proposed mechanisms, namely

\begin{itemize}[leftmargin=*]
\item \textbf{HiStGen$_{-P}$} removes the Markov paragraph dependency mechanism between context paragraphs, and the section boundary label is only decided by the representation of current paragraph.
\item \textbf{HiStGen$_{-S}$} removes the section-aware attention mechanism between a section and its heading. 
\item \textbf{HiStGen$_{-H}$} removes the Markov heading dependency mechanism between context headings, and the initial hidden state of the decoder is only decided by the representation of last paragraph in the section.
\item \textbf{HiStGen$_{-R}$} removes the review mechanism between context headings.
\item \textbf{HiStGen$_{-PSHR}$} removes all the mechanisms and reduces to a vanilla hierarchical sequence-to-sequence generation model.
\end{itemize}

\subsubsection{Step-wise Methods}

We also apply two types of step-wise process for the OG task.

\begin{itemize}[leftmargin=*]
\item \textbf{First-Identify-then-Generate} (\textbf{IG}). The first step is to identify the potential sections, and the second step is to generate the heading for each section. 
For the section identification step, based on the hidden representations of the input paragraphs (described in Section \ref{encoder}), we employ two methods: 
\begin{itemize}[leftmargin=*]
\item \textbf{Conditional random field} (\textbf{CRF}) is a well-known sequential labeling model. Here we follow the work \cite{huang2015bidirectional} where the CRF model is built upon the hierarchical encoder, and use the representation of the target paragraph and meanwhile take a chain dependence assumption between the labels, for section boundary prediction.
\item \textbf{Global paragraph dependency mechanism} (\textbf{GPD}) considers all the context paragraphs in a document, not just the previous and successive paragraph as in our Markov paragraph dependency mechanism, to predict the section boundary label for a target paragraph.
\end{itemize} 
For the heading generation step, we employ both extractive (TextRank and TopicRank) and generative (Hier and GHD) methods over the detected sections: 
\begin{itemize}[leftmargin=*]
\item \textbf{TextRank} \cite{mihalcea2004textrank} is a graph-based method inspired by the PageRank algorithm.
\item \textbf{TopicRank} \cite{bougouin2013topicrank} represents a document as a complete graph depending on a topical representation of the document.
\item \textbf{Hier} \cite{li2015hierarchical} takes the section as input using a hierarchical encoder structure (words form paragraph, paragraphs form section) and employs the section-aware attention (described in Section \ref{attention}) in the decoding phase.
\item \textbf{GHD} further employs a global heading dependency mech-
anism based on the Hier, where all the previous generated headings are taken into account to initialize the hidden state of the current decoder, not just the previous one as in our Markov heading dependency mechanism.
\end{itemize} 
By combining these two-step methods, we obtain eight types of IG methods denoted as \textbf{IG$_{\text{CRF+TextRank}}$}, \textbf{IG$_{\text{CRF+TopicRank}}$}, \textbf{IG$_{\text{CRF+Hier}}$}, \textbf{IG$_{\text{CRF+GHD}}$}, \textbf{IG$_{\text{GPD+TextRank}}$}, \textbf{IG$_{\text{GPD+TopicRank}}$}, \textbf{IG$_{\text{GPD+Hier}}$} and \textbf{IG$_{\text{GPD+GHD}}$}.

\item \textbf{First-Generate-then-Aggregate} (\textbf{GA}). The first step is to generate the heading for each paragraph, and the second step is to aggregate the paragraph with respect to their headings. For the heading generation step, we also employ the TextRank, TopicRank, Hier and GHD method over the paragraphs. For the heading aggregation step, we combine successive paragraphs with the same heading into one section. Similarly, we refer to these four types of GA process as \textbf{GA$_{\text{TextRank}}$}, \textbf{GA$_{\text{TopicRank}}$}, \textbf{GA$_{\text{Hier}}$} and \textbf{GA$_{\text{GHD}}$}.

\end{itemize}

\begin{table*}[t]
\centering
 \caption{Model analysis of our HiStGen model under the automatic evaluation. Two-tailed t-tests demonstrate the improvements of HiStGen to the variants are statistically significant ($^{\ddag}$ indicates $\text{p-value} < 0.01$).}
  \renewcommand{\arraystretch}{1.45}
   \setlength\tabcolsep{1.2pt}
  \begin{tabular}{c | c c c| c c c| c c c | c c c}  \hline \hline
   & \multicolumn{3}{c|}{celebrity} & \multicolumn{3}{c|}{cities} & \multicolumn{3}{c|}{music} & \multicolumn{3}{c}{mixture} \\ \hline

  Model & EM$_{outline}$ & EM$_{sec}$ & Rouge$_{head}$ &  EM$_{outline}$ & EM$_{sec}$ & Rouge$_{head}$ & EM$_{outline}$ & EM$_{sec}$ & Rouge$_{head}$ & EM$_{outline}$ & EM$_{sec}$ & Rouge$_{head}$ \\\hline 
  HiStGen$_{-P}$ & 13.95 & 32.81 & 58.36 & 16.82 & 38.54 & 57.01 & 12.65 & 30.06 & 58.05  & 13.62 & 32.25 & 57.69 \\
   HiStGen$_{-S}$ & 14.28 & 37.21 &  55.52 & 17.49 & 43.30 & 54.01 & 13.66 & 34.99 & 55.62 & 14.64 & 37.00 & 54.79\\
  HiStGen$_{-H}$ & 15.88 & 37.49 &  57.33 & 19.02 & 43.54 & 55.87 & 14.87 & 35.23 & 57.17 & 15.99 & 37.19 & 56.92\\ 
  HiStGen$_{-R}$ & 15.76 & 37.42 &  57.01 & 18.96 & 43.02 & 55.08 & 14.32   & 35.20 & 56.79 & 15.78  & 37.13 & 56.00\\ 
  HiStGen$_{-PSHR}$& 11.87 & 32.55 & 52.15 & 14.81 & 38.23 & 50.62 & 11.13 & 30.05 & 51.54 & 11.72 & 32.17 & 50.84\\ \hline
  HiStGen& $\textbf{17.07}^{\ddag}$ & \textbf{37.94} & \textbf{58.61} & $\textbf{20.34}^{\ddag}$ &  \textbf{43.81} & \textbf{57.26} & $\textbf{16.05}^{\ddag}$ & \textbf{35.66} &  \textbf{58.38} & $\textbf{16.74}^{\ddag}$ & \textbf{37.58} & \textbf{58.01} \\ \hline \hline
    \end{tabular}
   \label{table:ablation}
\end{table*}

\subsection{Evaluation Metrics}

To measure the quality of outline generated by our model and the baselines, we employ three automatic metrics, namely
\begin{itemize}[leftmargin=*]
\item \textbf{EM$_{outline}$}: evaluates the overall accuracy of the generated outline based on \textit{exact matching}. That is, if both the predicted section boundaries and the generated section headings in a document exactly match with the ground-truth, we treat the document as a positive sample. Otherwise the document is a negative sample. 
\item \textbf{EM$_{sec}$}: evaluates the accuracy of the section boundary prediction based on \textit{exact matching}. Namely, if the predicted section boundaries in a document exactly match with the ground-truth, we treat the document as a positive sample. Otherwise the document is a negative sample.
\item \textbf{Rouge$_{head}$} evaluates the similarities between generated headings and referenced headings only for the \textit{correctly predicted} sections. Specifically, we employ Rouge-1 \cite{lin2004rouge} to measure the uni-gram recall on the reference headings.
\end{itemize}

\subsection{Model Ablation}

We conduct ablation analysis to investigate the effect of proposed mechanisms in our HiStGen model. As shown in table  \ref{table:ablation}, we can observe that: 
(1) By removing the Markov paragraph dependence mechanism, the performance of $HiStGen_{-P}$ in terms of EM$_{sec}$ has a significant drop as compared with $HiStGen$. The results indicate that modeling the dependency between adjacent paragraphs does help decide the section boundaries. 
(2) $HiStGen_{-S}$ performs worse than $HiStGen_{-H}$ and $HiStGen_{-R}$ in terms of Rouge$_{head}$, showing that the coherence between a section and its heading (captured by the section-aware attention mechanism) has much bigger impact than that between context headings (captured by the Markov heading dependency mechanism and review mechanism) for heading generation. 
(3) HiStGen$_{PSHR}$ gives the worst performance, indicating that traditional seq2seq model without considering three-level coherence is not suitable for the OG task.
(4) By including all the mechanisms, $HiStGen$ achieves the best performance in terms of all the evaluation metrics.

\begin{table}[t]
\centering
 \caption{Comparisons between our HiStGen and step-wise baselines in terms of EM$_{outline}$(\%).}
  \renewcommand{\arraystretch}{1.45}
   \setlength\tabcolsep{7pt}
  \begin{tabular}{c | c | c | c  | c}  \hline \hline
   & celebrity & cities & music & mixture \\ \hline 
  IG$_{\text{CRF+TextRank}}$  & 2.82 & 2.49 & 2.25 & 2.76   \\
   IG$_{\text{CRF+TopicRank}}$ & 2.23 & 2.03 & 2.15 & 2.68 \\
   IG$_{\text{CRF+Hier}}$ & 14.36 & 16.27 & 12.74 & 13.06  \\  
   IG$_{\text{CRF+GHD}}$ & 14.58 & 16.86 & 12.90 & 13.12  \\ \hline
   IG$_{\text{GPD+TextRank}}$ & 2.19 & 1.87 & 1.66 & 1.95  \\
   IG$_{\text{GPD+TopicRank}}$ & 2.01 & 1.34 & 1.26 & 1.80  \\
   IG$_{\text{GPD+Hier}}$ & 10.88 & 12.34 & 11.11 & 11.20  \\
   IG$_{\text{GPD+GHD}}$ & 11.18 & 12.78 & 11.30 & 11.54 \\\hline 
   GA$_{\text{TextRank}}$ & 0.93 & 0.87 & 0.11 & 0.24  \\ 
   GA$_{\text{TopicRank}}$ & 0.22 & 0.43 & 0.02 & 0.16  \\
   GA$_{\text{Hier}}$ & 4.16 & 4.09 & 3.97 & 3.99 \\ 
   GA$_{\text{GHD}}$ & 4.56 & 4.79 & 4.41 & 4.42 \\  
\hline
 HiStGen& \textbf{17.07} & \textbf{20.34} & \textbf{16.05} & \textbf{16.74}\\ \hline
     \hline
    \end{tabular}
   \label{table:overall}
\end{table}

\begin{figure*}[t]
	\centering
		\includegraphics[scale=0.385]{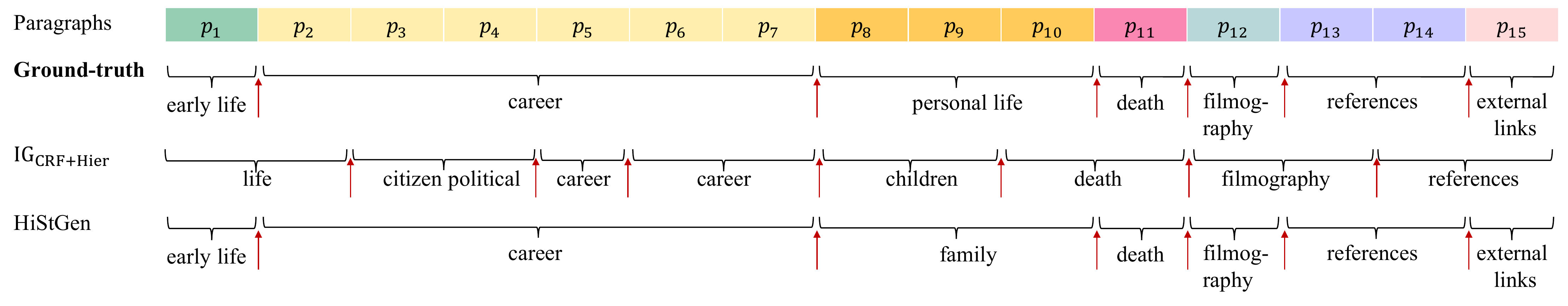}
		\caption{An example from the test W{\scriptsize IKI}OG data. $p_1$ to $p_{15}$ are the paragraphs in the article. Red colored arrows stand for the section boundaries, and texts below brackets stand for the section headings. The two results below are the outputs of the IG$_{CRF+Hier}$ and HiStGen model.} 
\label{fig:case}
\end{figure*}

\subsection{Baseline Comparison}

\subsubsection{Overall performance}

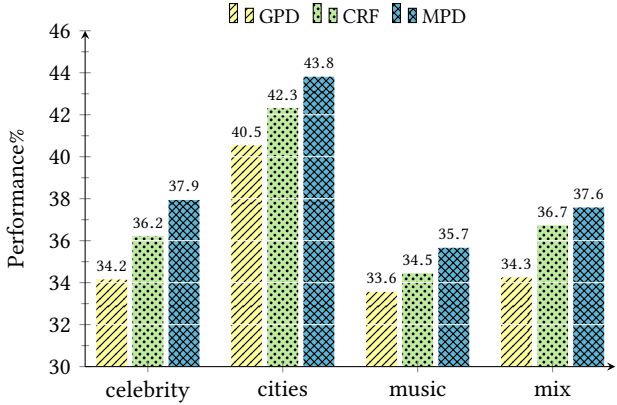
\begin{figure}[t]
\centering 
\begin{tikzpicture}
 \begin{axis}[
  ybar,
  axis on top,
        height=.34\textwidth,
        width=0.485\textwidth,
        bar width=0.41cm,
        ytick={30,32,34,36,38,40,42,44,46},
        ymajorgrids, tick align=inside,
        major grid style={draw=white},
        minor y tick num={1},
        enlarge y limits={value=.1,upper},
        ymin=30, ymax=46,
        axis lines=left,
        enlarge x limits=0.155,
        legend style={
            at={(0.5,1.1)},
            font=\small,
            anchor=north,
            draw=none,
            legend columns=-1,
            /tikz/every even column/.append style={column sep=0.1cm}
        },
        ylabel={Performance\%},
        ylabel style={
            anchor=south,
            at={(ticklabel* cs:0.5)},
            yshift=-15pt
        },
        symbolic x coords={
           celebrity,cities,music,mix},
       xtick=data,
       nodes near coords={
        \pgfmathprintnumber[fixed zerofill,precision=1]{\pgfplotspointmeta}
       },
       every node near coord/.append style={anchor=south, font=\fontsize{6pt}{4pt}\selectfont},
 ]
 \addplot [draw=none, fill=c1, postaction={pattern=north east lines}] coordinates {
      (celebrity,34.15)
      (cities, 40.54)
      (music,33.56)
      (mix,34.25)
      };
   \addplot [draw=none,fill=c2, postaction={pattern=crosshatch dots}] coordinates {
      (celebrity,36.21)
      (cities, 42.31)
      (music,34.45)
       (mix,36.73)
      };
   \addplot [draw=none, fill=c3, postaction={pattern= crosshatch}] coordinates {
      (celebrity,37.94)
      (cities, 43.81)
      (music,35.66)
      (mix, 37.58)
      };
      \legend{GPD, CRF, MPD}
 \end{axis}

\end{tikzpicture}
\caption{Performance comparison of the section boundary prediction under EM$_{sec}$ metric.}
\label{fig:section}
\end{figure}

The overall performance comparisons between our HiStGen and the step-wise baselines are shown in Table \ref{table:overall}. We have the following observations: (1) The $GA$ process (i.e., $GA_{\text{TextRank}}$, $GA_{\text{TopicRank}}$, $GA_{\text{Hier}}$ and $GA_{\text{GHD}}$) performs very poorly. By looking at the results of the $GA$ methods, we find that $GA$ tends to segment the document into too much sections since it usually generates different headings even for paragraphs that should belong to a same section.
(2) For the $IG$ process, the methods based on $CRF$ perform better than that based on $GPD$. For example, the relative improvement of $IG_{\text{CRF+GHD}}$ over $IG_{\text{GPD+GHD}}$ is about $13.7\%$ in terms of EM$_{outline}$ on the mixture set. We analyze the results and find that using $CRF$ can obtain better section prediction results, showing that the dependency on the context labels is more important than that on all the paragraphs for section identification. Moreover, for the $IG$ process, the generative methods can achieve significantly better results than the extractive methods, since those extractive methods are unsupervised in nature. 
(3) Our $HiStGen$ model can outperform all the step-wise baselines significantly (p-value $<$ 0.01). As compared with the best-performing baseline $IG_{\text{CRF+GHD}}$, the relative improvement of $HiStGen$ over $IG_{\text{CRF+GHD}}$ is about $27.6\%$ in terms of EM$_{outline}$ on the mixture set. The results demonstrate the effectiveness of our end-to-end learning model.

\subsubsection{Section prediction performance}

We further compare the section boundary prediction performance between our Markov paragraph dependency mechanism (\textbf{MPD} for short) and the two baseline methods, i.e., $CRF$ and $GPD$, by keeping the rest components the same. The results are shown in Figure \ref{fig:section}. We can find that: (1) The improvements of $MPD$ over $GPD$, showing that the consideration of the previous and successive paragraph is better than the consideration of all the paragraphs in a document for section boundary prediction. The reason might be by considering all the paragraphs, $GPD$ tends to bring noisy information that may hurt the prediction on section boundaries. Moreover, $GPD$ leads to much higher computing complexity than $MPD$ (i.e., $O(M^2)>O(2M)$). 
(2) $MPD$ performs better than $CRF$, demonstrating that depending on the semantic representations of the previous and successive paragraph is more beneficial than only depending on the labels of the previous and successive paragraph in section boundary prediction. All the improvements over the baselines are statistically significant (p-value < 0.01).

\subsubsection{Section heading prediction performance}

We evaluate the section heading generation ability to demonstrate the effectiveness of our Markov heading dependency mechanism and review mechanism. Here we suppose that sections in an article are already given, and only need to predict the corresponding headings for each section. We consider two generative baselines $Hier$ and $GHD$, where $GHD$ is an extension of $Hier$ by employing a global heading dependency mechanism. We then introduce our Markov heading dependency mechanism based on the $Hier$, named \textbf{Hier$_{MHD}$}, and further employ the review mechanism, named \textbf{Hier$_{MHD+RE}$}. All these methods employ the section-aware attention in generation. The performance under Rouge$_{head}$ is shown in Table \ref{table:heading}. We can find that: (1) \textit{Hier} performs worst among all the methods, showing that the independence between context headings is not good for section heading generation. 
(2) By incorporating all the previous generated headings to model the dependence between context headings, $GHD$ shows slight improvements on the heading generation performance. It indicates that the global dependency may not be effective in heading generation by involving too much context information, and also leads to high computing complexity.
(3) The improvements of $Hier_{MHD}$ over $GHD$ indicate that the dependency between adjacent headings is sufficient for generating good and consistent section headings.
(4) The improvements of $Hier_{MHD+RE}$ over $Hier_{MHD}$ demonstrate that the review mechanism is also helpful in improving the quality of section heading generation. All the improvements over the baselines are statistically significant (p-value $<$ 0.01).

\begin{table}[t]

\centering
 \caption{Evaluation results(\%) of the section heading generation under Rouge$_{head}$ metric when the real sections are given aforehead.}
  \renewcommand{\arraystretch}{1.25}
   \setlength\tabcolsep{8pt}
  \begin{tabular}{c | c | c | c | c }  \hline \hline
   & celebrity & cities & music & mixture \\ \hline
  Hier & 57.54 & 57.16  & 58.31  & 57.81 \\ 
  GHD & 57.75 & 57.74  & 58.52  & 57.93 \\ \hline
   Hier$_{MHD}$  & 58.96  & 59.46  & 58.78 & 58.99  \\
   Hier$_{MHD+RE}$  & \textbf{59.43}  & \textbf{59.94}  & \textbf{59.22} &  \textbf{59.37}  \\
     \hline \hline
    \end{tabular}
   \label{table:heading}
\end{table}

\subsection{Case Study}

To better understand how different models perform, we conduct some case studies. We take one Wikipedia article\footnote{\url{https://en.wikipedia.org/wiki/Sabu_Dastagir}} from the ``celebrity'' test data as an example. As shown in Figure \ref{fig:case}, there are $15$ paragraphs in this article, which are segmented into $7$ sections. We show the identified sections and generated headings from our model as well as that from the baseline model $IG_{\text{CRF+Hier}}$. We can find that: (1) The number of sections predicted by $IG_{\text{CRF+Hier}}$ is larger than the ground-truth (i.e., $8>7$) and the segmentation is totally wrong. The results show that using current paragraph representation and context label dependency, CRF may not be able to  make correct section boundary prediction. (2) Without considering the coherence between context headings, $IG_{\text{CRF+Hier}}$ generates repetitive headings (e.g., ``career'' repeats twice) and the heading with inconsistent style (e.g., ``citizen political'' is not suitable for the description of a celebrity). (3) Our $HiStGen$ can generate right section boundaries and consistent headings. Note that $HiStGen$ generates ``family'' for the third section whose true heading is ``personal life''. As we look at that section, we found that ``family'' is actually a very proper heading and $HiStGen$ did not generate the ``personal life'' as the heading possibly due to the review mechanism by avoiding partial duplication with the ``early life'' heading.

\section{Conclusion and future work}
In this paper we introduced a challenging OG task to unveil the inherent content structure of a multi-paragraph document by identifying its potential sections and generating the corresponding section headings. 
To tackle the problem, we formulated the OG task as a hierarchical structured prediction problem and developed a novel hierarchical structured neural generation model to capture the three levels of coherence.
Furthermore, we built a new benchmark dataset W{\scriptsize IKI}OG to study and evaluate the OG task. The experimental results demonstrated that our model can well capture the inherent content structure of documents. In the future work, we would like to extend our model to produce hierarchical outlines for documents.

\section{Acknowledgments}

This work was funded by the National Natural Science Foundation of China (NSFC) under Grants No. 61425016, 61722211, 61773362, and 61872338, the Youth Innovation Promotion Association CAS under Grants No. 20144310, and 2016102, the National Key R\&D Program of China under Grants No. 2016QY02D0405, and the Foundation and Frontier Research Key Program of Chongqing Science and Technology Commission (No. cstc2017jcyjBX0059).

\bibliographystyle{ACM-Reference-Format}
\bibliography{sample-bibliography}

\end{document}